\documentclass{article}

   \usepackage[preprint]{neurips_2020}

\usepackage[utf8]{inputenc} % allow utf-8 input
\usepackage[T1]{fontenc}    % use 8-bit T1 fonts
\usepackage{hyperref}       % hyperlinks
\usepackage{url}            % simple URL typesetting
\usepackage{booktabs}       % professional-quality tables
\usepackage{amsfonts}       % blackboard math symbols
\usepackage{nicefrac}       % compact symbols for 1/2, etc.
\usepackage{microtype}      % microtypography

\usepackage{multirow}
\usepackage{subfigure}
\usepackage{graphicx}
\usepackage{amsmath}

\title{Finet: Using Fine-grained Batch Normalization to Train Light-weight Neural Networks}

\author{
  Chunjie Luo\\
  Institute of Computing Technology\\
  Chinese Academy of Sciences\\
  \texttt{luochunjie@ict.ac.cn} \\
\And
  Jianfeng Zhan\\
  Institute of Computing Technology\\
  Chinese Academy of Sciences\\
  \texttt{zhanjianfeng@ict.ac.cn} \\
\And
  Lei Wang\\
  Institute of Computing Technology\\
  Chinese Academy of Sciences\\
  \texttt{wanglei\_2011@ict.ac.cn} \\
\And
  Wanling Gao\\
  Institute of Computing Technology\\
  Chinese Academy of Sciences\\
  \texttt{gaowanling@ict.ac.cn} \\
}

\begin{document}

\maketitle

\begin{abstract}
To build light-weight network, we propose a new normalization, Fine-grained Batch Normalization (FBN). Different from Batch Normalization (BN), which normalizes the final summation of the weighted inputs, FBN normalizes the intermediate state of the summation. We propose a novel light-weight network based on FBN, called Finet. At training time, the convolutional layer with FBN can be seen as an inverted bottleneck mechanism. FBN can be fused into convolution at inference time. After fusion, Finet uses the standard convolution with equal channel width, thus makes the inference more efficient. On ImageNet classification dataset, Finet achieves the state-of-art performance (65.706\% accuracy with 43M FLOPs, and 73.786\% accuracy with 303M FLOPs), Moreover, experiments show that Finet is more efficient than other state-of-art light-weight networks. 
\end{abstract}

\section{Introduction}

Since AlexNet \cite{Krizhevsky2012ImageNetCW} won ImageNet Large-Scale Visual Recognition Competition in 2012, deep neural networks have received great successes in many areas of machine intelligence. Modern state-of-art networks \cite{Simonyan2015VeryDC} \cite{He2016DeepRL} \cite{Huang2017DenselyCC} \cite{Szegedy2016RethinkingTI} \cite{Zoph2018LearningTA} become deeper and wider. The requirement of high computational resources hinders their usages on many mobile and embedded applications. As a result, there has been rising interest for the design of light-weight neural networks \cite{Howard2017MobileNetsEC} \cite{Sandler2018MobileNetV2IR} \cite{Howard2019SearchingFM} \cite{Tan2018MnasNetPN} \cite{Zhang2018ShuffleNetAE} \cite{Sandler2018MobileNetV2IR} \cite{Huang2017CondenseNetAE} \cite{Zhang2017InterleavedGC} \cite{Xie2018IGCV2IS} \cite{Sun2018IGCV3IL}. The objective of light-weight networks is to decrease the computation complexity with low or no loss of accuracy.

\begin{figure}[!htb]
\centering
\subfigure[BN]{
\label{fig-bn}
\includegraphics[scale=0.22]{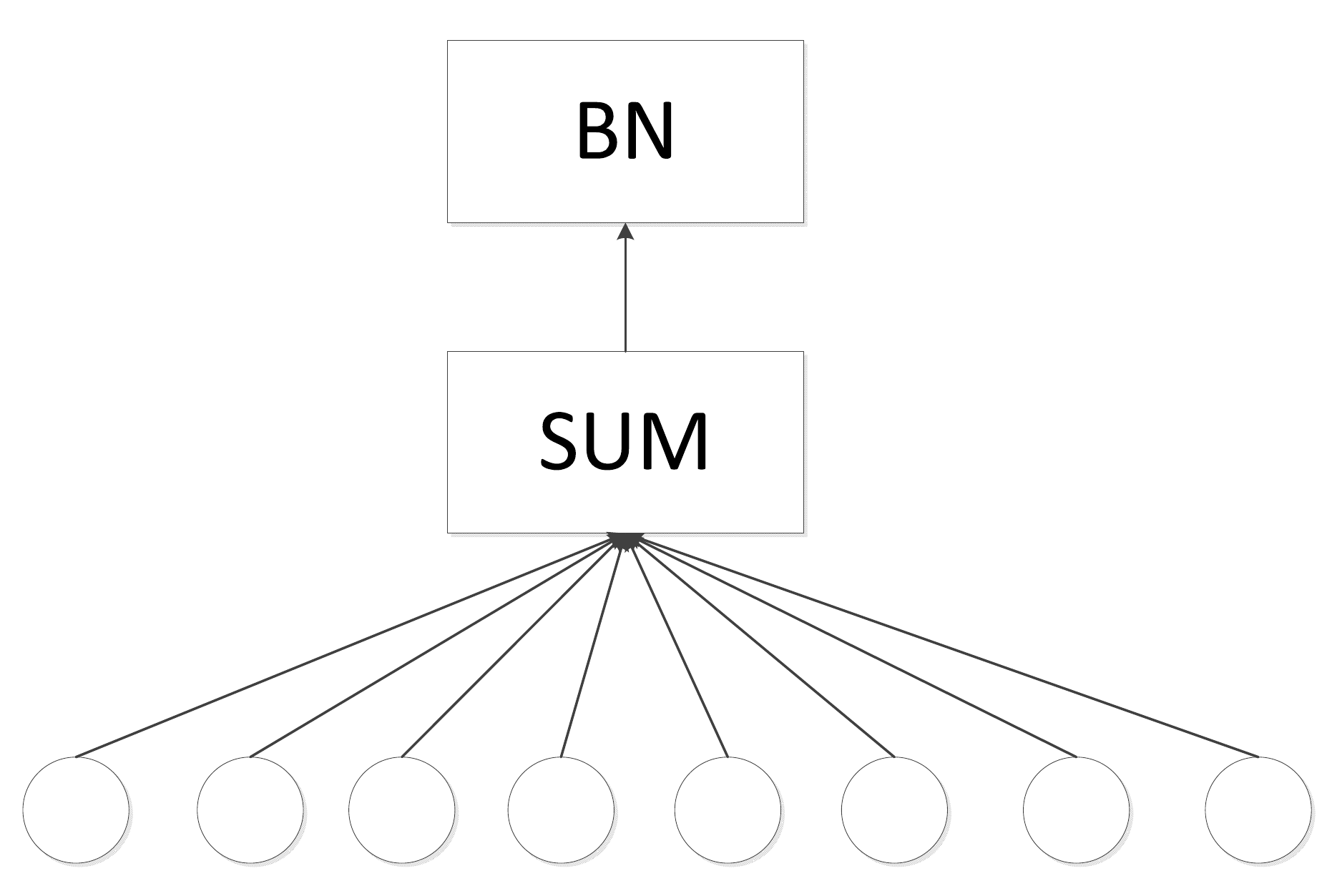}}
\centering
\subfigure[BN]{
\label{fig-bn2}
\includegraphics[scale=0.22]{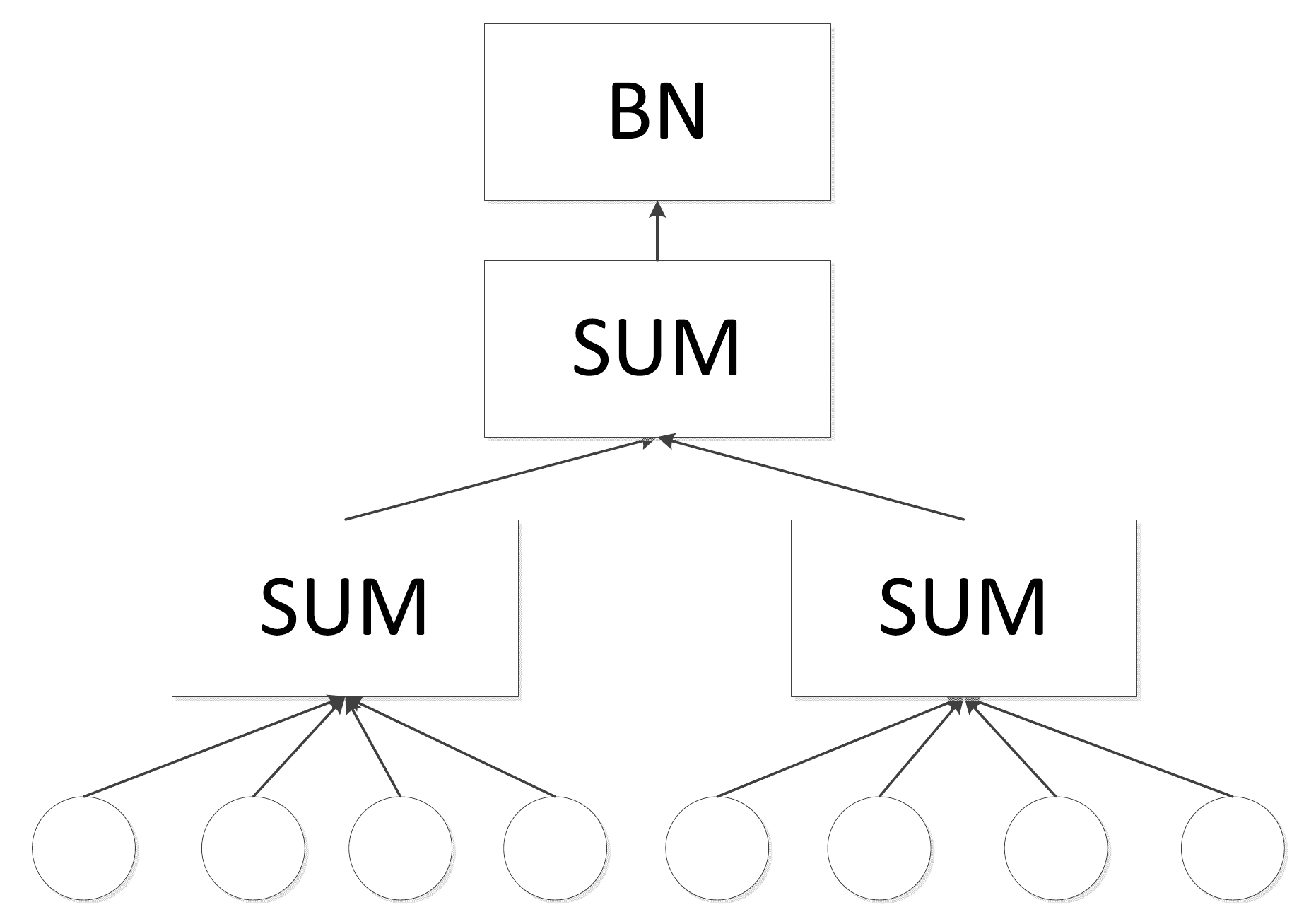}}
\centering
\subfigure[FBN]{
\label{fig-fbn}
\includegraphics[scale=0.22]{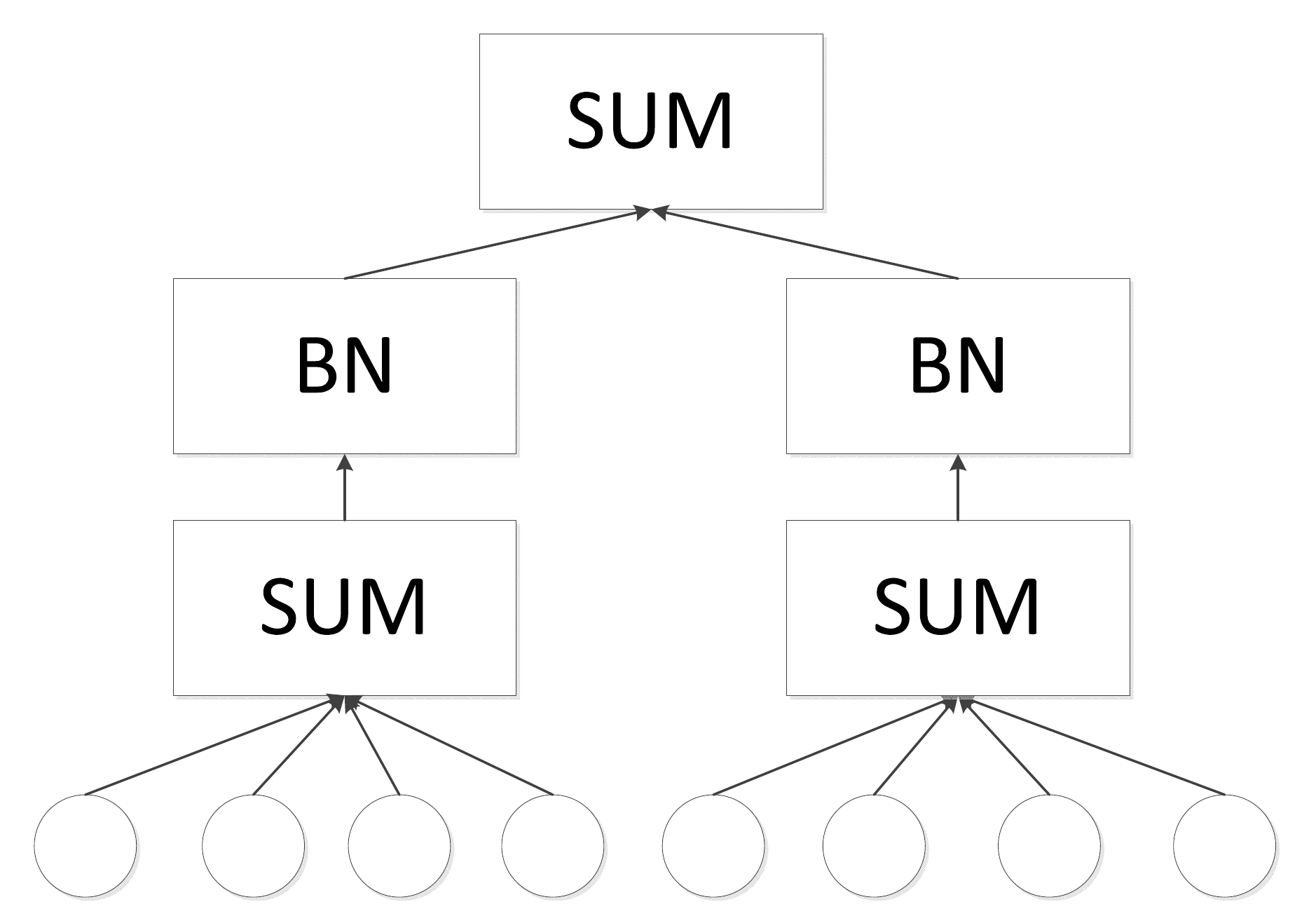}}
\caption{The difference between Batch Normalization (BN) and Fine-grained Batch Normalization (FBN). Each small circle in the figure represents $w_{i} x_{i}$. Mathematically, (a) is equal to (b) because of Associative Law of addition. (c) represents FBN which normalizes the intermediate state of the summation. At inference time, the normalization can be fused into the linear transformation $w_{i} x_{i}$. After fusion, (a), (b) and (c) are equivalent.}
\label{fig-diff}
\end{figure}

To build light-weight networks, we propose a new normalization, Fine-grained Batch Normalization (FBN). Batch Normalization (BN) \cite{ioffe2015batch} has become a standard component in modern neural networks. There are lots of variants of Batch Normalization to meet diverse usage scenarios \cite{ioffe2017batch} \cite{ba2016layer} \cite{ulyanov2016instance} \cite{wu2018group}  \cite{Luo2020ExtendedBN} . FBN aims to improve the training of light-weight networks, while keep the inference efficient. As shown in Figure \ref{fig-diff}, different from BN, which normalizes the final summation of the weighted inputs, FBN normalizes the intermediate state of the summation. At inference time, the normalization can be fused into the linear transformation. Then there is no need for the intermediate state.

We propose a novel light-weight network based on FBN, called Finet. At training time, the convolutional layer with FBN can be seen as an inverted bottleneck mechanism since the intermediate channels are normalized and then summarized. However, this bottleneck has only one convolutional layer. FBN can be fused into convolution at inference time. After fusion, Finet uses the standard convolution with equal channel width, thus makes the inference more efficient.

On ImageNet classification dataset, Finet achieves the state-of-art performance. With 43M FLOPs, Finet achieves 65.706\% accuracy, outperforming the corresponding model of ShuffleNetV2 (60.3\%), MobileNetV2 (58.2\%), MobileNetV3 (65.4\%). With 303M FLOPs, Finet achieves 73.786\% accuracy, outperforming the corresponding model of ShuffleNetV2 (72.6\%), MobileNetV2 (72.0\%). Moreover, we compare the inference speed of Finet, ShuffleNetV2, MobileNetV2, MobileNetV3, and MnasNet on three different mobile phones. The results show that Finet is more efficient than other state-of-art light-weight networks. 
We also evaluate Finet on CIFAR-10/100 dataset, and present the influences of different hyper-parameter settings. Finally, we show that FBN also improves the performance of ResNet.

\section{Related Work}
\subsection{Normalization}

Batch Normalization \cite{ioffe2015batch} performs the normalization for each training minibatch along (N,H,W) dimensions in the case of NCHW format feature. Normalization Propagation \cite{arpit2016normalization} uses a data-independent parametric estimate of the mean and standard deviation instead of explicitly calculating from data. 
Batch Renormalization \cite{ioffe2017batch} introduces two extra parameters to correct the fact that the minibatch statistics differ from the population ones. 
Layer Normalization \cite{ba2016layer} computes the mean and standard deviation along (C,H,W) dimensions. 
Instance Normalization \cite{ulyanov2016instance} computes the mean and standard deviation along (H,W) dimensions. 
Group Normalization \cite{wu2018group} is a intermediate state between layer normalization and instance normalization. 
Extended Batch Normalization \cite{Luo2020ExtendedBN} computes the mean along the (N, H, W) dimensions, and computes the standard deviation along the (N, C, H, W) dimensions.
Weight Normalization \cite{salimans2016weight} normalizes the filter weights instead of the activations by re-parameterizing the incoming weight vector.  
Cosine Normalization \cite{luo2017cosine} normalizes both the filter weights and the activations by using cosine similarity or Pearson correlation coefficient instead of dot product in neural networks.
Kalman Normalization \cite{wang2018kalman} estimates the mean and standard deviation of a certain layer by considering the distributions of all its preceding layers.
Instead of the standard $L^{2}$ Batch Normalization, \cite{hoffer2018norm} performs the normalization in $L^{1}$ and $L^{\infty}$ spaces. Generalized Batch Normalization \cite{yuan2019generalized} investigates a variety of alternative deviation measures for scaling and alternative mean measures for centering.
Batch-Instance Normalization \cite{nam2018batch} uses a learnable gate parameter to combine batch and instance normalization together, and Switchable Normalization \cite{luo2018differentiable}  uses learnable parameters to combine batch, instance and layer normalization.
Virtual Batch Normalization \cite{salimans2016improved} and spectral normalization \cite{miyato2018spectral} focus on the normalization in generative adversarial networks. 
Self-Normalizing \cite{klambauer2017self} focuses on the fully-connected networks.
Recurrent Batch Normalization \cite{cooijmans2016recurrent} modifies batch normalization to use in recurrent networks. 
EvalNorm \cite{singh2019evalnorm} estimates corrected normalization statistics to use for batch normalization during evaluation. 
\cite{ren2016normalizing} provides a unifying view of the different normalization approaches. \cite{santurkar2018does}, \cite{luo2018towards} and \cite{bjorck2018understanding} try to explain how Batch Normalization works.

\subsection{Light-weight Network}

MobileNet \cite{Howard2017MobileNetsEC} is a light-weight deep neural network designed and optimized for mobile and embedded vision applications. MobileNet is based on depthwise separable convolutions to reduce the number of parameters and computation FLOPs. MobileNetV2 \cite{Sandler2018MobileNetV2IR} introduces two optimized mechanisms: 1) inverted residual structure where the shortcut connections are between the thin layers. 2) linear bottlenecks which removes non-linearities in the narrow layers. MnasNet \cite{Tan2018MnasNetPN} is a neural architecture automated searched for mobile device by using multi-objective optimization and factorized hierarchical search space. MobileNetV3 \cite{Howard2019SearchingFM} is also a light-weight network searched by network architecture search algorithm. EfficientNet \cite{Tan2019EfficientNetRM} proposes a new scaling method that uniformly scales all dimensions of depth/width/resolution
ShufflfleNet \cite{Zhang2018ShuffleNetAE} utilizes pointwise group convolution and channel shuffle, to greatly reduce computation cost while maintaining accuracy.
ShufflfleNetV2 \cite{Sandler2018MobileNetV2IR} proposes to evaluate the direct metric on the target platform, beyond only considering FLOPs. Following several practical guidelines, a new efficient architecture based on channel split and shuffle is presented.
CondenseNet \cite{Huang2017CondenseNetAE} combines dense connectivity with learned group convolution. IGCV \cite{Zhang2017InterleavedGC} \cite{Xie2018IGCV2IS} \cite{Sun2018IGCV3IL} propose interleaved group convolutions to build efficient networks. SqueezeNet \cite{Iandola2017SqueezeNetAA} aims to decrease the number of parameters while maintaining competitive accuracy.

\section{Fine-grained Batch Normalization}
\subsection{Batch Normalization}
Batch normalization (BN) has become a standard technique for training the deep networks. 
For a neuron, BN is defined as

\begin{equation} \label{eq-bn}
\widehat{x} = (\sum_{i=1}^{N}{w_{i}x_{i}}  - \mu ) / \sigma
\end{equation}

where $w_i$ is the input weight, and $x_i$ is the input, $N$ is the size of input, $\mu$ is the mean of the neurons along the batch dimension, and $\sigma$ is the standard deviation. For simplicity, we omit the epsilon in the denominator, and the affine transformation after normalization. The non-linear function $f(\widehat{x})$ is performed after the normalization. 

\subsection{Fine-grained Batch Normalization}

We can divide the inputs and the weights of a neuron into $G$ groups. Because of Associative Law of addition, we have 

\begin{equation} \label{eq-add}
\sum_{i=1}^{N}{w_{i}x_{i}} = \sum_{g=1}^{G}\sum_{j=1}^{N/G}{w_{gj}x_{gj}}
\end{equation}

where $w_{i}$ and $w_{gj}$ are the same weights, $x_{i}$ and $x_{gj}$ are the same inputs, but using different index notation. Then Equation \ref{eq-bn} can be re-written as  

\begin{equation} \label{eq-bn2}
\widehat{x} = (\sum_{g=1}^{G}\sum_{j=1}^{N/G}{w_{gj}x_{gj}}  - \mu ) / \sigma
\end{equation}

In this paper, we propose Fine-grained Batch Normalization (FBN), which is defined as

\begin{equation} \label{eq-fbn}
\widehat{x} = \sum_{g=1}^{G}  ((\sum_{j=1}^{N/G}{w_{gj}x_{gj}}  - \mu_g)  / \sigma_g)
\end{equation}

where $\mu_g$ and $\sigma_g$ are the mean and the standard deviation of the intermediate summation along the batch dimension.
Different from BN, which normalizes the final summation of the weighted inputs, FBN normalizes the intermediate state of the summation. When training with BN, neurons can be coordinated in a mini-batch.  FBN makes the coordination more fine-grained. 
Figure \ref{fig-diff} shows the difference between BN and FBN.  
At training time, the intermediate state need to be stored and traced. Thus FBN takes more memory resource for training. Though FBN needs to normalize more channels, the computation overhead is trivial comparing to the convolution operation.

\subsection{Inference and Normalization Fusion}

At inference time, the mean and the standard deviation are pre-computed from the training data by the moving average. The inference of BN is computed as 

\begin{equation} \label{eq-bn-infer}
\widehat{x} = (\sum_{i=1}^{N}{w_{i}x_{i}}  - \mu^{,}) / \sigma^{,}
\end{equation}

where $\mu^{,}$ refers to the moving average of the mean,  $\sigma^{,}$  refers to the moving average of the standard deviation. Since the mean and the standard deviation are pre-computed and fixed at inference time, the normalization can be fused into the linear transformation, e.g. convolution operation. 
Re-write Equation \ref{eq-bn-infer} as

\begin{equation} \label{eq-bn-infer2}
\widehat{x} = \sum_{i=1}^{N}{\frac{w_{i}}{\sigma^{,}}x_{i}}  - \frac{\mu^{,}} {\sigma^{,}}
\end{equation}

Let $w_{i}^{,}=\frac{w_{i}}{\sigma^{,}}$, and $b^{,}=\frac{\mu^{,}} {\sigma^{,}}$, then we can fuse the normalization into the linear transformation as

\begin{equation} \label{eq-bn-fusion}
\widehat{x} = \sum_{i=1}^{N}{w_{i}^{,}x_{i}}  - b^{,}
\end{equation}

Here $w_{i}^{,}$ is the new weight of the linear transformation,  and $b^{,}$ is the bias.  Thus BN is fused into the new linear transformation. Similarly, the inference of FBN is computed as

\begin{equation} \label{eq-fbn-infer}
\begin{aligned}
\widehat{x}   &=\sum_{g=1}^{G}  ((\sum_{j=1}^{N/G}{w_{gj}x_{gj}}  - \mu_g^{,})  / \sigma_g^{,}) \\
                    &= \sum_{g=1}^{G}  (\sum_{j=1}^{N/G}{ \frac{w_{gj}}{\sigma_g^{,}} x_{gj}}  - \frac{\mu_g^{,}}{ \sigma_g^{,}}) \\
                   &=\sum_{g=1}^{G}  \sum_{j=1}^{N/G}{ \frac{w_{gj}}{\sigma_g^{,}} x_{gj}}  - \sum_{g=1}^{G}   \frac{\mu_g^{,}}{ \sigma_g^{,}}
\end{aligned}
\end{equation}

where $\mu_g^{,}$ and $\sigma_g^{,}$ are the moving average of the mean and the standard deviation of intermediate summation. 
Let $w_{gj}^{,}=\frac{w_{gj}}{\sigma_g^{,}}$, and $b^{,}=\sum_{g=1}^{G}   \frac{\mu_g^{,}}{ \sigma_g^{,}}$. Because of Equation \ref{eq-add}, FBN can also be fused into the linear transformation as

\begin{equation} \label{eq-fbn-fuse}
\begin{aligned}
\widehat{x} &= \sum_{g=1}^{G}  \sum_{j=1}^{N/G}{ w_{gj}^{,} x_{gj}}  - b^{,} \\
                   &= \sum_{i=1}^{N} w_{i}^{,} x_{i} - b^{,}
\end{aligned}
\end{equation}

In summary, we can fuse FBN into the linear transformation at inference time. By this way, we do not need to store the intermediate state of the summation. 
%For example, we just need the standard convolutional operation. There is no need for group convolution and channel expansion any more. 
That is to say, there is no computation and memory overhead at inference time.

\section{Network Architecture}

\begin{figure}[!htb]
\centering
\subfigure[Stride=1]{
\label{fig-block1}
\includegraphics[scale=0.3]{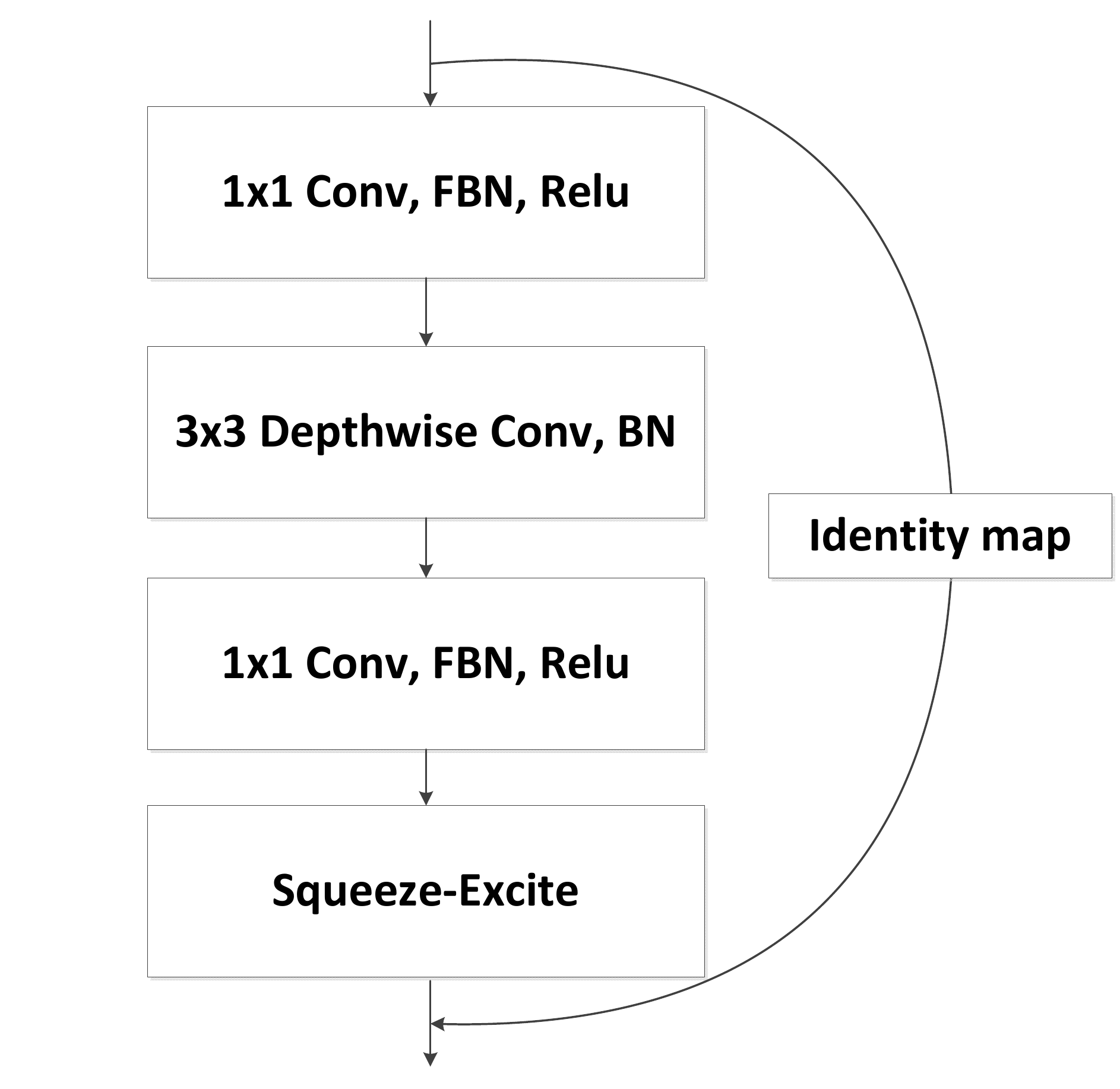}}
\centering
\subfigure[Stride=2]{
\label{fig-block2}
\includegraphics[scale=0.3]{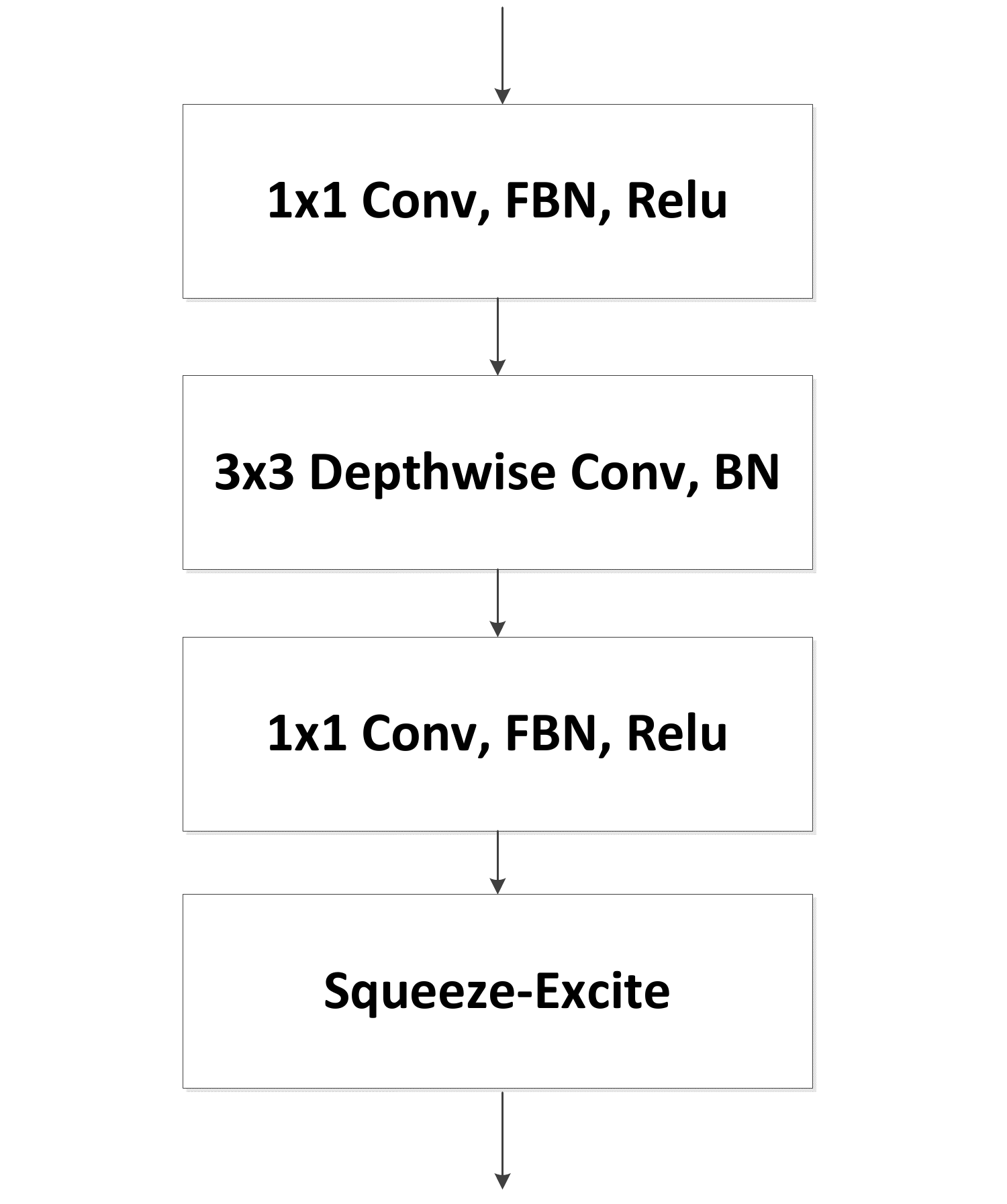}}
\caption{The building blocks of Finet}
\label{fig-block}
\end{figure}

\begin{figure}[!htb]
\centering
\subfigure[]{
\label{fig-imple1}
\includegraphics[scale=0.3]{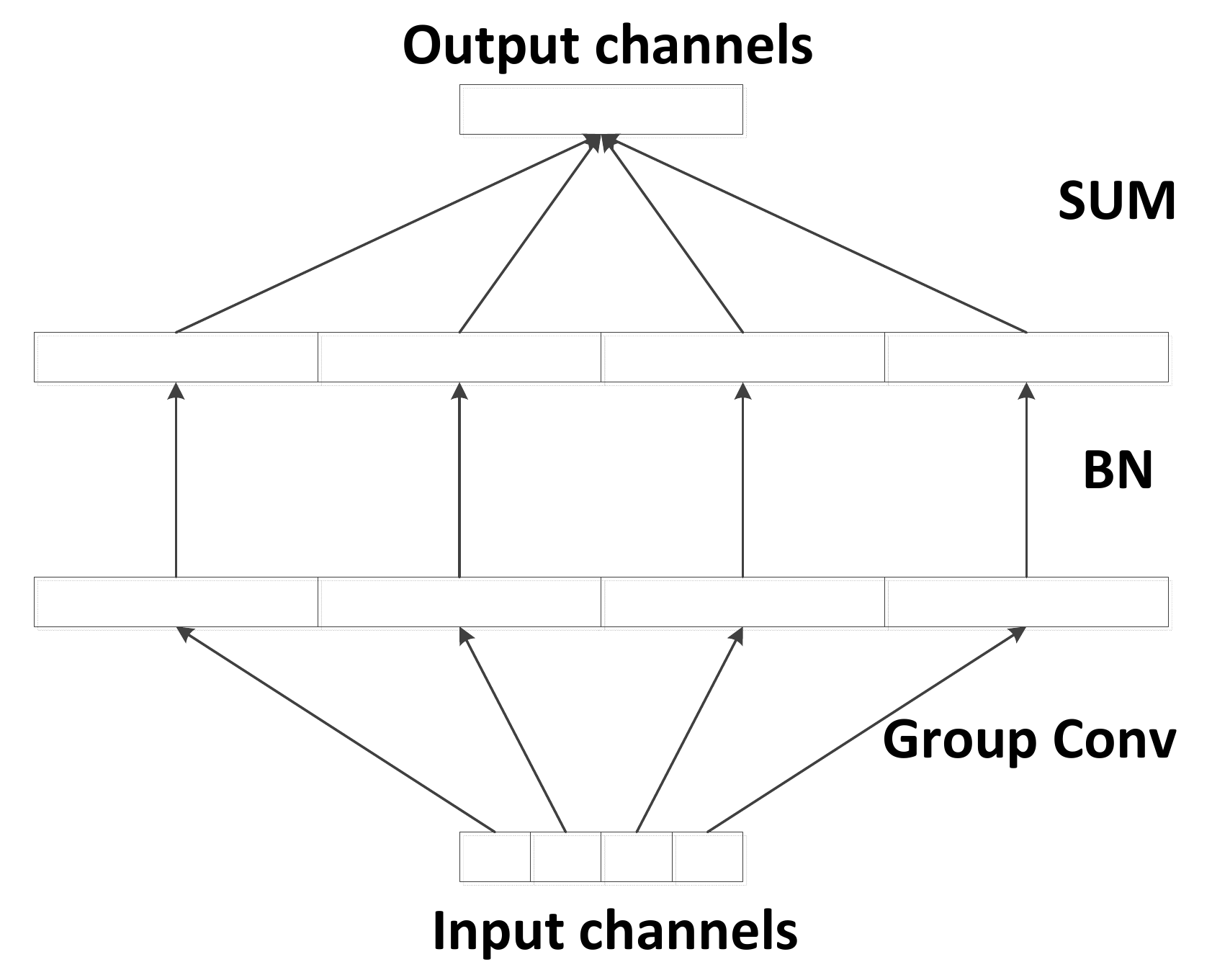}}
\centering
\subfigure[]{
\label{fig-imple2}
\includegraphics[scale=0.3]{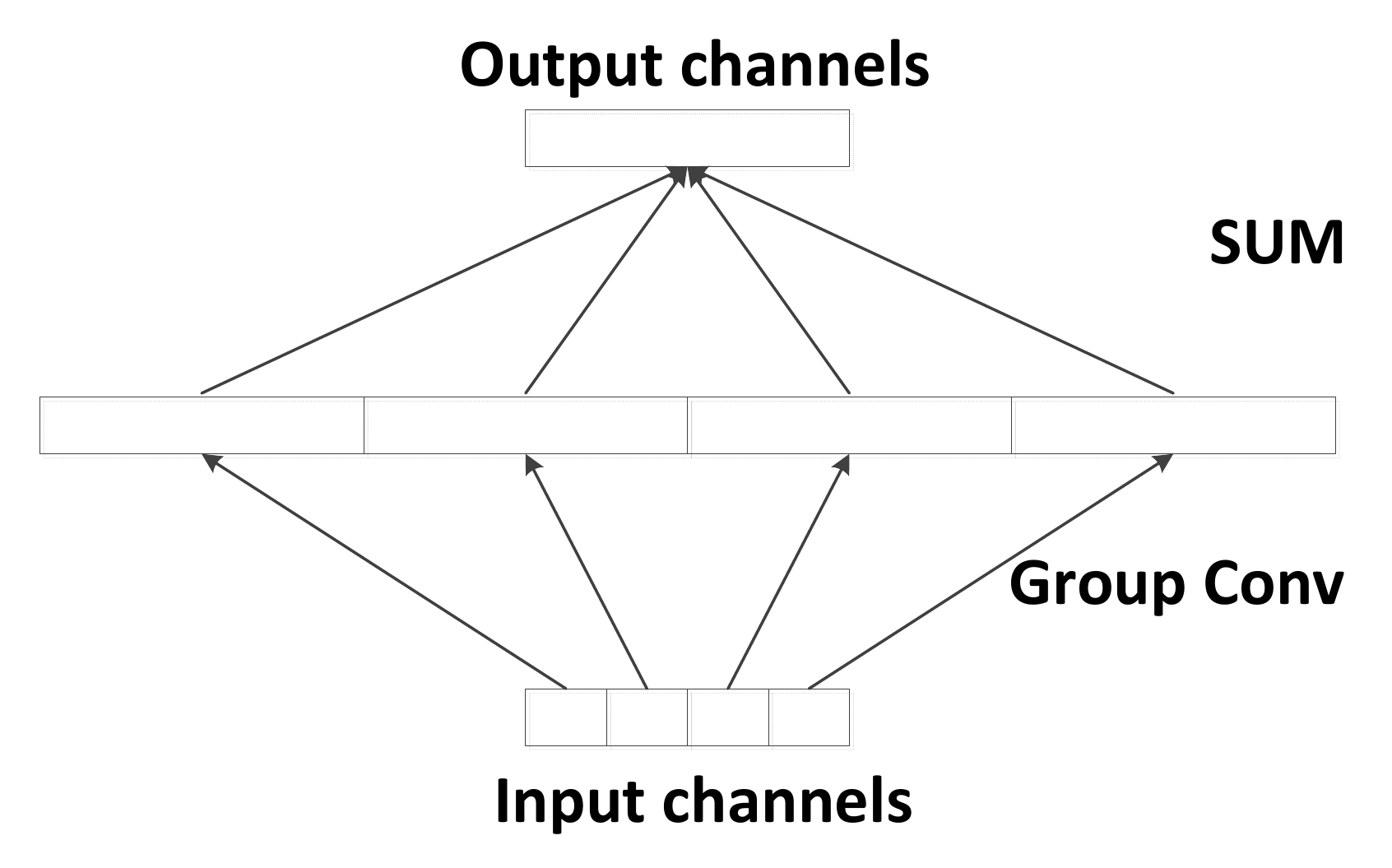}}
\centering
\subfigure[]{
\label{fig-imple1}
\includegraphics[scale=0.3]{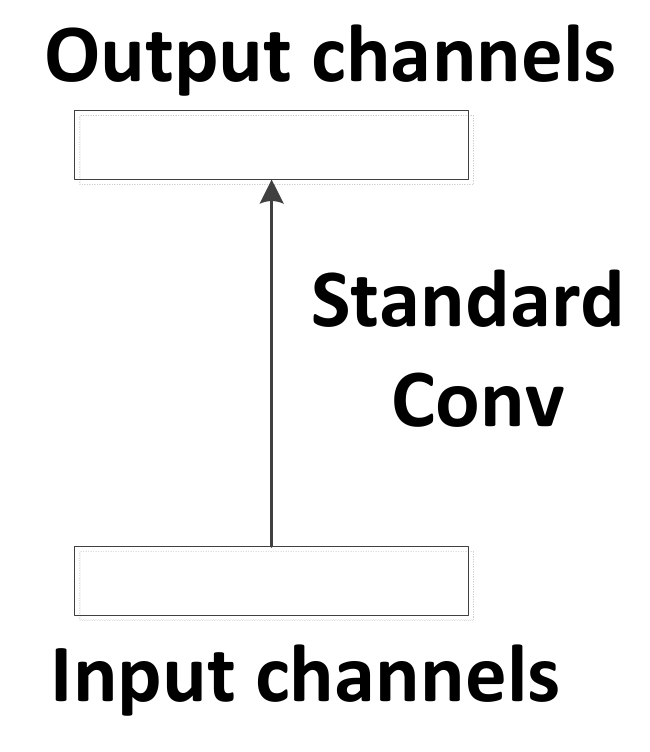}}
\caption{The implementation of Fine-grained Batch Normalization. (a) FBN can be seen as an inverted bottleneck mechanism at training time.  (b) Normalization can be fused into convolution at inference time. (c) The standard convolution can be used at inference time. (b) and (c) are equal.}
\label{fig-FBN-implement}
\end{figure}

In this section, we describe the architecture of our light-weight network based on Fine-grained Batch Normalization, called Finet.  Figure \ref{fig-block} shows the building blocks of Finet. Similar as other light-weight networks, we use the depthwise convolution to reduce the FLOPs and the parameters.  The main difference of our block is that we use FBN in the 1x1 pointwise convolution instead of BN.
As pointed in \cite{Zagoruyko2016WideRN} \cite{Zhang2018ShuffleNetAE} \cite{Zhang2017InterleavedGC}, wider layer makes more powerful representation, but brings more FLOPs and parameters. There are many ways to reduce the FLOPs and parameters of wide layer, e.g. bottleneck \cite{He2016DeepRL} or inverted bottleneck \cite{Sandler2018MobileNetV2IR}, group convolution \cite{Zhang2018ShuffleNetAE}. 
%Bottleneck can factorize a complex convolution (NxN) into two simple convolution (NxM+MxN), while group convolution makes the connection more sparse. 
Finet normalizes the intermediate channels, which are $G$ times wider than final channels.  
Take a convolutional layer as example, as shown in Figure \ref{fig-FBN-implement}, we can use group convolution to implement FBN. The channels are expanded by $G$ times, then the expanded channels are normalized and summarized.  It can be seen as an inverted bottleneck mechanism. However, this bottleneck has only one convolutional layer. 
As pointed in \cite{Ma2018ShuffleNetVP}, non-equal channel width and group convolution increase memory access cost. Fortunately, FBN can be fused into convolution at inference time. After fusion, Finet uses the standard convolution with equal channel width, thus makes the inference more efficient. That is very helpful for deploy on the mobile and embedded device which has limit computation and memory resource, or online service environment which is sensitive to the latency. 

%\begin{figure}[!htb]
%\centering
%\label{fig-FBN-implement}
%\includegraphics[scale=0.3]{figs/FBN-implement2}
%\caption{The implementation of Fine-grained Batch Normalization. It can be seen as an inverted bottleneck mechanism. At inference time, normalization can be fused into convolution, and then the standard convolution can be used.}
%\end{figure}

FBN can also be treated as a procedure of splitting, transforming, and aggregating, which is the key design philosophy of ResNeXt \cite{Xie2017AggregatedRT}. Different from ResNeXt which uses multi-layer network as the transforming, we use normalization as the transforming. Comparing to ResNeXt, we do not need to carry out splitting, transforming, and aggregating at inference time because of normalization fusion.

We can add Squeeze-Excite module \cite{Hu2018SqueezeandExcitationN} which is a light-weight attention mechanism and widely used in other light-weight networks, e.g. Mnasnet \cite{Tan2018MnasNetPN}, MobileNetV3 \cite{Howard2019SearchingFM}. For simplicity, we do not use identity map in the block where the stride of depthwise convolution is 2.  

Table \ref{table-arch} shows the overall architecture of Finet, for small and large levels of complexities. We adopt the architecture similar with ShufflenetV2. The architectures of the series of MobileNet are more heterogeneous, thus bring difficulties for optimization of performance and memory usage. For Finet, there are two differences from ShufflenetV2: 1) the building blocks shown in Figure \ref{fig-block} are used in each stage, 2) there is an additional fully connected layer before the classifier layer. 
The fully connected layer is prone to overfitting since it takes much parameters. As a result, modern heavy-weight convolutional networks \cite{Lin2014NetworkIN} \cite{He2016DeepRL} \cite{Huang2017DenselyCC} \cite{Szegedy2016RethinkingTI} try to avoid using fully connected layer except the final classifier layer. However, light-weight networks usually suffer from underfitting rather than overfitting. For light-weight networks, fully connected layer brings much parameters but with little computation overhead. The latest light-weight network MobilenetV3 \cite{Howard2019SearchingFM} also uses an additional fully connected layer to redesign expensive layers. Squeeze-Excite module also utilizes the fully connected layer to enhance the representative ability with little computation overhead. 

\begin{table}[!htb]
  \caption{Overall architecture of Finet, for small and large levels of complexities. The architecture is similar with ShufflenetV2, except 1) the building blocks of Finet are used in each stage, 2) there is an additional fully connected layer before the classifier layer. }
  \label{table-arch}
  \centering
  \begin{tabular}{|c|c|c|c|c|c|c|}
   \hline
    \multirow{2}{*}{Layer} & \multirow{2}{*}{Output size} & \multirow{2}{*}{Ksize }& \multirow{2}{*}{Stride} & \multirow{2}{*}{Repeat }& \multicolumn{2}{c|}{Output channels }\\
    \cline{6-7}
     & & & & & Small & Large \\
    \hline
    Image & 224x224 & & & &3 &3 \\
   \hline
   Conv1 & 112x112 & 3x3  & 2 &  \multirow{2}{*}{1} & \multirow{2}{*}{24} & \multirow{2}{*}{24}\\
   MaxPool & 56x56 & 3x3 & 2&  & & \\
   \hline
   \multirow{2}{*}{Stage2} & 28x28 & & 2 & 1 & \multirow{2}{*}{30} & \multirow{2}{*}{100} \\
    & 28x28 & & 1 & 3 &&\\ 
    \hline
   \multirow{2}{*}{Stage3} & 14x14 & & 2 & 1 & \multirow{2}{*}{60} & \multirow{2}{*}{200} \\
    & 14x14 & & 1 & 7 &&\\ 
    \hline
   \multirow{2}{*}{Stage4} & 7x7 & & 2 & 1 & \multirow{2}{*}{120} & \multirow{2}{*}{400} \\
    & 7x7 & & 1 & 3 &&\\ 
    \hline
   Conv5 & 7x7 &1x1&1 &1& 1024& 1024 \\
    \hline
   GlobalPool & 1x1 & 7x7 & & 1&1024&1024 \\
   \hline
   FC1 & 1x1 & & & 1& 1024&1024  \\
   \hline
   FC2 & 1x1 & & & 1& 1000&1000  \\
   \hline
%   \hline
%   Paras (w/ SE) &&&&& 40M& 300M \\
%  \hline
%   Paras (w/o SE) &&&&& 40M& 300M \\
%  \hline
%    FLOPs &&&&& 40M& 300M \\
%   \hline
  \end{tabular}
\end{table}

\section{Experiment}

\subsection{ImageNet}
ImageNet classification dataset \cite{russakovsky2015imagenet} has 1.28M training images and 50,000 validation images with 1000 classes.  We use Pytorch in our experiments. To augment data, we use the same procedure as the official examples of Pytorch \cite{pytorch}.
The training images are cropped with random size of 0.08 to 1.0 of the original size and a random aspect ratio of 3/4 to 4/3 of the original aspect ratio, and then resized to 224x224. Then random horizontal flipping is made. The validation image is resized to 256x256, and then cropped by 224x224 at the center. Each channel of the input is normalized into 0 mean and 1 std globally.
We use SGD with 0.9 momentum,  and 4e-5 weight decay. Four TITAN Xp GPUs are used to train the networks. The batch size is set to 512. We use linear-decay learning rate policy (decreased from 0.2 to 0). Dropout with 0.2 is used in the last two fully connected layers. We train the networks with 320 epochs. For Squeeze-Excite module, we set the number of hidden unit to 200.

\begin{figure}[!htb]
\centering
\includegraphics[scale=0.4]{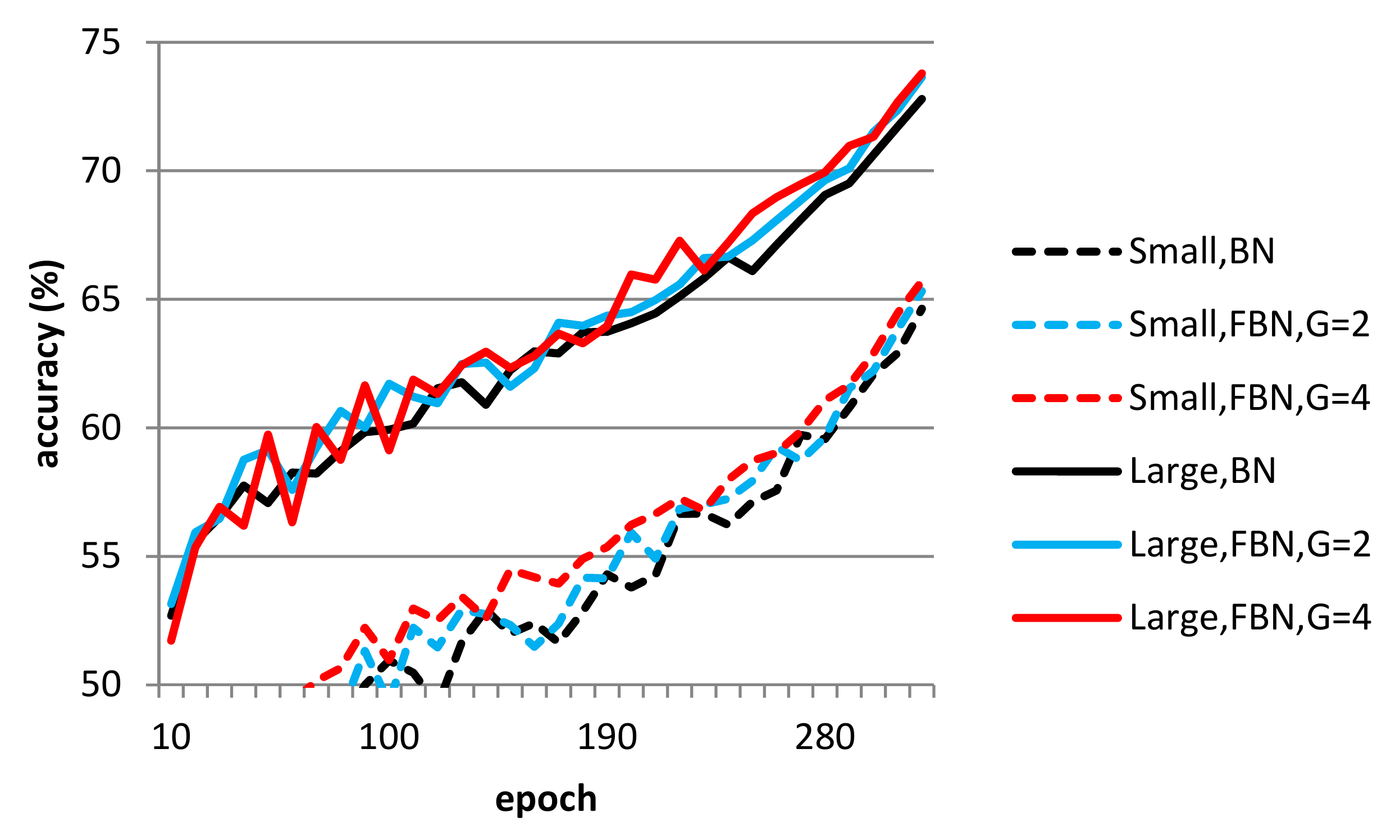}
\caption{The validation accuracy of Finet with Squeeze-Excite, vs. numbers of training epochs}
\label{fig-imagenet}
\end{figure}

Figure \ref{fig-imagenet} and Table \ref{table-imagenet} show the results of ImageNet classification. The baseline network uses BN. Actually, BN is equal to FBN when $G=1$. As G increases, Finet achieves higher accuracy with little overhead of FLOPs and parameters. Moreover, Squeeze-Excite module enhances the accuracy. With Squeeze-Excite and $G=4$, small Finet achieves 65.706\% accuracy with 43M FLOPs, and large Finet achieves 73.786\% accuracy with 303M FLOPs. Finet outperforms ShuffleNetV2 (60.3\% with 41M FLOPs, and 72.6\% with 299M FLOPs) and MobileNetV2 (58.2\% with 43M FLOPs, and 72.0\% with 300M FLOPs) which are also manual designed architectures. 
Finet takes more parameters than ShuffleNetV2 and MobileNetV2. The last two fully connected layers take about 2M parameters. 
Small version of Finet achieves higher accuracy than the corresponding model of MobileNetV3, while
MobileNetV3 and MnasNet achieve higher accuracies for the large version. MobileNetV3 and MnasNet are auto-searched architectures. Neural Architecture Search (NAS) reduces the demand for experienced human experts comparing to hand-drafted design. It can find the optimal combination of existing technique units, but can not invent new techniques. Moreover, the searched architectures are often more heterogeneous, and are less efficient than the homogeneous architectures, as shown in the next Section \ref{sec-speed}. 

\begin{table}[!htb]
  \caption{The FLOPs, Parameters (Paras), and Accuracies (Acc) of different networks on ImageNet validation data. BN: Batch Normalization, FBN: Fine-grained Batch Normalization, SE: Squeeze-Excite.}
  \label{table-imagenet}
  \centering
  \begin{tabular}{|c|c|c|c|c|c|c|}
   \hline
   & \multicolumn{3}{c|}{Small} & \multicolumn{3}{c|}{Large}\\
   \cline{2-7}
   &  FLOPs & Paras & Acc & FLOPs & Paras & Acc \\
   &  (Million) & (Million) & (\%) & (Million) & (Million) & (\%) \\
   \hline
   baseline  (BN) & 42.6 & 2.388 & 61.710& 301.8&4.434&71.256\\
   Finet (FBN,G=2) & 42.6 & 2.392& 62.988& 301.8&4.448& 72.410\\
   Finet (FBN,G=4) & 42.6 & 2.402& 63.338 &301.8& 4.477& 72.746\\
   \hline
   baseline+SE (BN) & 43.0&2.809& 64.632 &303.2&5.812& 72.790\\
   Finet+SE (FBN,G=2) & 43.0&2.813& 65.322&303.2&5.827&73.642 \\
   Finet+SE (FBN,G=4) & 43.0&2.827 & \textbf{65.706} &303.2&5.855& \textbf{73.786} \\
   \hline
   \hline
  ShuffleNetV2 \cite{Ma2018ShuffleNetVP} & 41 & 1.4& 60.3&299& 3.5& 72.6\\
  MobileNetV2 \cite{Sandler2018MobileNetV2IR} \cite{slim}& 43 & 1.66 & 58.2&300&3.4&72.0\\
  \hline
  MobileNetV3 \cite{Howard2019SearchingFM}& 44 & 2.0& 65.4& 219& 5.4& \textbf{75.2}\\
  MnasNet \cite{Tan2018MnasNetPN}& - & - & - & 312& 3.9 & \textbf{75.2}\\
  \hline
  \end{tabular}
\end{table}

\subsection{Inference Speed} \label{sec-speed}

We evaluate the inference speed of different light-weight networks on three mobile phones: Samsung Galaxy s10e, Huawei Honor v20, and Vivo x27. Galaxy s10e and Vivo x27 equip the mobile SoC of Qualcomm SnapDragon. SnapDragon uses a heterogeneous computing architecture to accelerate the AI applications. AI Engine of SnapDragon consists of Kryo CPU cores, Adreno GPU and Hexagon DSP.
Honor v20 equips HiSilicon Kirin SoC. Different from SnapDragon, Kirin introduces a specialized neural processing unit (NPU) to accelerate the AI applications. The detailed configurations of the devices are shown in Table \ref{table-device}. 

\begin{table}[tbh!]
\caption{The configurations of the measured devices}
\centering
\begin{tabular}{|c|c|c|c|c|c|}
\hline
Device &  Soc & CPU  & AI Accelerator  & RAM \\
\hline
\multirow{2}{*}{Galaxy s10e} & \multirow{2}{*}{Snapdragon 855} &  \multirow{2}{*}{Kryo 485, 2.84 GHz, 7nm}   & Adreno 640 GPU  & \multirow{2}{*}{6GB}   \\
& &  & Hexagon 690 DSP &  \\
\hline
Honor v20  & Kirin 980 & Cortex-A76, 2.6 GHz, 7nm & Cambricon NPU  & 8GB  \\
\hline
\multirow{2}{*}{Vivo x27}  &  \multirow{2}{*}{Snapdragon 710}  & \multirow{2}{*}{Kryo 360, 2.2 GHz, 10nm} &  Adreno 616 GPU &  \multirow{2}{*}{8GB}  \\
& &  & Hexagon 685 DSP &  \\
\hline
  \end{tabular}
  \label{table-device}
\end{table}

We use Pytorch Mobile \cite{pymobile} to deploy the networks on mobile phones. We compare the large versions of Finet, ShufflenetV2, MobileNetV2, MobileNetV3, and MnasNet. Normalization are omitted since it can be fused into convolution operation at inference time. Each model infers 1000 validation images sequentially, and the average throughput is counted. Table \ref{table-speed} shows the results of different models. On all of the three phones, Finet without Squeeze-Excite module is the fastest model, and the second fastest model is Finet with Squeeze-Excite module. The heterogeneous models (MobileNetV2, MobileNetV3, and MnasNet) are slower than homogeneous models (Finet, ShufflenetV2)

\begin{table}[!htb]
  \caption{Inference speed (images per second)}
  \label{table-speed}
  \centering
  \begin{tabular}{|c|c|c|c|}
   \hline
                 & Galaxy s10e & Honor v20 & Vivo x27\\
\hline
Finet, without SE & \textbf{7.80} &	\textbf{6.34}	& \textbf{2.45} \\
Finet, with SE &  \textbf{7.69}  & \textbf{6.17} & \textbf{2.37} \\
ShufflenetV2 & 7.35 	& 5.99 & 2.09 \\
MobileNetV2 &  3.98  & 3.65 & 1.42 \\
MobileNetV3 &  5.65 &	4.82 & 1.79 \\
MnasNet &  3.77  & 3.74 & 1.48 \\

\hline
  \end{tabular}
\end{table}

\subsection{CIFAR}
CIFAR-10 \cite{krizhevsky2009learning} is a dataset of natural 32x32 RGB images in 10 classes with 50, 000 images for training and 10, 000 for testing. CIFAR-100 is similar with CIFAR-10 but with 100 classes. To augment data, the training images are padded with 0 to 36x36 and then randomly cropped to 32x32 pixels. Then randomly horizontal flipping is made. Each channel of the input is normalized into 0 mean and 1 std globally.
Large Finet is evaluated in this section. To adapt Finet to CIFAR datasets, the stride of Conv1 is set to 1, and omit the MaxPool in Table \ref{table-arch}.
We use SGD with 0.9 momentum, and 5e-4 weight decay. All models are trained on one TITAN Xp GPU. The batch size is set to 128. Learning rate is set to 0.1, and decreased 10 times at epoch 100 and 150. We train the networks with 200 epochs. For Squeeze-Excite module, we set the number of hidden unit to 200.

Table \ref{table-diffG} shows the accuracies of Finet on CIFAR-10/100 with different groups.
FBN achieves better accuracy than BN (when $G=1$, FBN is equal to BN).
When $G=8$, Finet achieves the highest accuracy of 93.674\% on CIFAR-10, and 77.202\% on CIFAR-100.
The affine transformation in normalization is define as $y = \gamma \widehat{x} + \beta$, where $\gamma$ and $\beta$ are learned parameters for each channel in convolutional networks. Since the intermediate channels of FBN are wider than BN, there are more parameters in the affine transformation. To analyze how FBN improve the performance, we also evaluate FBN without affine transformation. In that case, there is no extra parameter comparing to BN. As shown in Table \ref{table-diffG}, FBN still achieves better accuracy than BN without affine transformation. Table \ref{table-diffC} shows the accuracies of Finet on CIFAR-10/100 by fixing the number of input channels per group. Because the layers have different channel numbers, the group number $G$ changes across layers in this setting. Generally, less input channels per group make more fine-grained training and increase the accuracy.

\begin{table}[!htb]
  \caption{The accuracies (\%) of Finet on CIFAR-10/100 with different groups}
  \label{table-diffG}
  \centering
  \begin{tabular}{|c|c|c|c|c|c|c|}
   \hline
                & & G=1 & G=2 & G=4 & G=6 & G=8\\
\hline
\multirow{2}{*}{CIFAR-10}  &with affine &   92.584 &	93.082&	93.532&	93.286	& \textbf{93.674}  \\
&  without affine &     91.958 &	93.162&	92.478&	92.564&	\textbf{93.430}   \\
\hline
\multirow{2}{*}{CIFAR-100}  &with affine &   75.624 &	76.752&	76.856&	76.716	& \textbf{77.202}  \\
&  without affine &   74.216   & 74.382 &	74.486&	\textbf{74.772}&	74.734   \\
\hline
  \end{tabular}
\end{table}

\begin{table}[!htb]
  \caption{The accuracies (\%) of Finet on CIFAR-10/100 with different input channels per group}
  \label{table-diffC}
  \centering
  \begin{tabular}{|c|c|c|c|}
   \hline
                 & C/G=20 & C/G=50 & C/G=100\\
  \hline
CIFAR-10 &    93.470&	\textbf{93.606} &	93.122\\
\hline
CIFAR-100 &   \textbf{76.926} &	76.850 &	75.674\\
%without affine transformation &         &         &          \\
\hline
  \end{tabular}
\end{table}

Finally, we evaluate ResNet18 and ResNet50 with FBN on CIFAR datasets. The batch size is set to 64 since training ResNet with FBN consumes a lot of GPU memories. Even decreasing the batch size, there is no enough memory for training ResNet50 with $G=4$ on single TITAN xp. Thus we only evaluate ResNet50 with $G=2$. Table \ref{table-resnet} shows the accuracies of Resnet18 and ResNet50 with FBN on CIFAR-10/100.  
The results show that FBN also achieves better accuracy than BN ($G=1$) for training heavy-weight networks.

\begin{table}[!htb]
  \caption{The accuracies (\%) of ResNet with FBN on CIFAR-10/100}
  \label{table-resnet}
  \centering
  \begin{tabular}{|c|c|c|c|c|}
   \hline
              &   & G=1 & G=2 & G=4\\
  \hline
\multirow{2}{*}{CIFAR-10} &ResNet18 &    94.512&	\textbf{95.012}&	94.962\\
& ResNet50 &      94.644 &	\textbf{95.040} & - \\
\hline
\multirow{2}{*}{CIFAR-100} &ResNet18 &   76.866&	\textbf{77.754}&	77.726\\
& ResNet50 &      77.796 &	\textbf{78.122} & - \\
\hline
  \end{tabular}
\end{table}

\section{Conclusion}

In this paper, we propose a new normalization, Fine-grained Batch Normalization (FBN), and a novel light-weight network based on FBN, called Finet. At training time, the convolutional layer with FBN can be seen as an inverted bottleneck mechanism. At inference time, Finet uses the standard convolution with equal channel width after normalization fusion, thus makes the inference more efficient.
We show the effectiveness and efficiency of Finet in our experiments.

%\section*{Broader Impact}

%\clearpage
\bibliographystyle{plain}
\bibliography{ref}

\end{document}